\documentclass[letterpaper, 10 pt, conference]{ieeeconf} 

\IEEEoverridecommandlockouts                              

\usepackage{float}
\usepackage{lipsum} 
\usepackage{multicol} 
\usepackage{cite}
\usepackage{soul,color}
\usepackage{amsmath,amssymb,amsfonts}
\usepackage{algorithm}
\usepackage{algpseudocode}
\usepackage{graphicx}
\usepackage{textcomp}
\usepackage{xcolor}
\usepackage{hyperref} 
\usepackage[affil-it]{authblk}
\let\labelindent\relax
\usepackage{enumitem}

\usepackage{booktabs}
\usepackage{colortbl}
\usepackage{booktabs} 
\usepackage{multirow}

\newcommand{\Method}[1]{\State \hspace{-0.35cm}{\textbf{Method:}} #1}
\newcommand{\Input}[1]{\State \hspace{-0.35cm}{\textbf{Input:}} #1}
\newcommand{\Output}[1]{\State \hspace{-0.35cm}{\textbf{Output:}} #1}

\definecolor{citecolor}{RGB}{192, 57, 43} 
\definecolor{linkcolor}{RGB}{39, 174, 96} 
\definecolor{urlcolor}{RGB}{41, 128, 185} %

\hypersetup{
    colorlinks=true,
    linkcolor=linkcolor, 
    citecolor=citecolor, 
    urlcolor=urlcolor 
}

\begin{document}

\title{\LARGE \bf \texttt{MERLION}$\vcenter{\hbox{\includegraphics[scale=0.08]{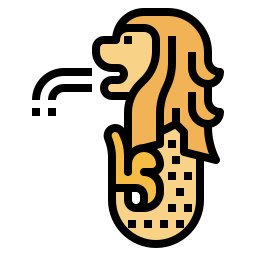}}}$: \textbf{M}arine \textbf{E}xplo\textbf{R}ation with \textbf{L}anguage gu\textbf{I}ded \textbf{O}nline i\textbf{N}formative Visual Sampling and Enhancement}
\author{Shrutika Vishal Thengane\text{*}\thanks{\text{*} These authors contributed equally.}, Marcel Bartholomeus Prasetyo\text{*}, Yu Xiang Tan, Malika Meghjani
    \thanks{The authors are with Singapore University of Technology and Design (SUTD), Singapore.
    {\tt\small \{shrutika\_thengane, marcel\_prasetyo,  malika\_meghjani\}@sutd.edu.sg}, {\tt\small yuxiang\_tan@mymail.sutd.edu.sg}}
}

\maketitle


\begin{abstract}

Autonomous and targeted underwater visual monitoring and exploration using Autonomous Underwater Vehicles (AUVs) can be a challenging task due to both online and offline constraints. The online constraints comprise limited onboard storage capacity and communication bandwidth to the surface, whereas the offline constraints entail the time and effort required for the selection of desired keyframes from the video data. An example use case of targeted underwater visual monitoring is finding the most interesting visual frames of fish in a long sequence of an AUV's visual experience. This challenge of targeted informative sampling is further aggravated in murky waters with poor visibility. In this paper, we present MERLION, a novel framework that provides semantically aligned and visually enhanced summaries for murky underwater marine environment monitoring and exploration. Specifically, our framework integrates (a) an image-text model for semantically aligning the visual samples to the user's needs, (b) an image enhancement model for murky water visual data and (c) an informative sampler for summarizing the monitoring experience. We validate our proposed MERLION framework on real-world data with user studies and present qualitative and quantitative results using our evaluation metric and show improved results compared to the state-of-the-art approaches. 
We have open-sourced the code for MERLION at the following link \href{https://github.com/MARVL-Lab/MERLION.git}{https://github.com/MARVL-Lab/MERLION.git}.

\end{abstract}


\section{Introduction}
Vision-based solutions used for autonomous underwater exploration, data collection and monitoring tasks are the most non-invasive and convenient for deployment on Autonomous Underwater Vehicles (AUVs). However, the quality of the visual data can be highly compromised in deep or murky waters due to poor lighting conditions and presence of particulates. In addition, depending on the desired task and duration of the underwater exploration, there could be strict or demanding requirements for on-board data storage or communication bandwidth for sending the data in real time to the remote user. Compromised visual data is solvable with visual data enhancement while limited communication bandwidth could be solved with visual sampling. Visual sampling refers to making an optimal choice for a representative subset of a long-horizon visual feed. In this paper, we address both of these challenges for the application of long-term vision-based exploration and monitoring in murky waters.

\begin{figure}[t]
    \centering
    \includegraphics[width=\linewidth, trim=50 70 70 70, clip]{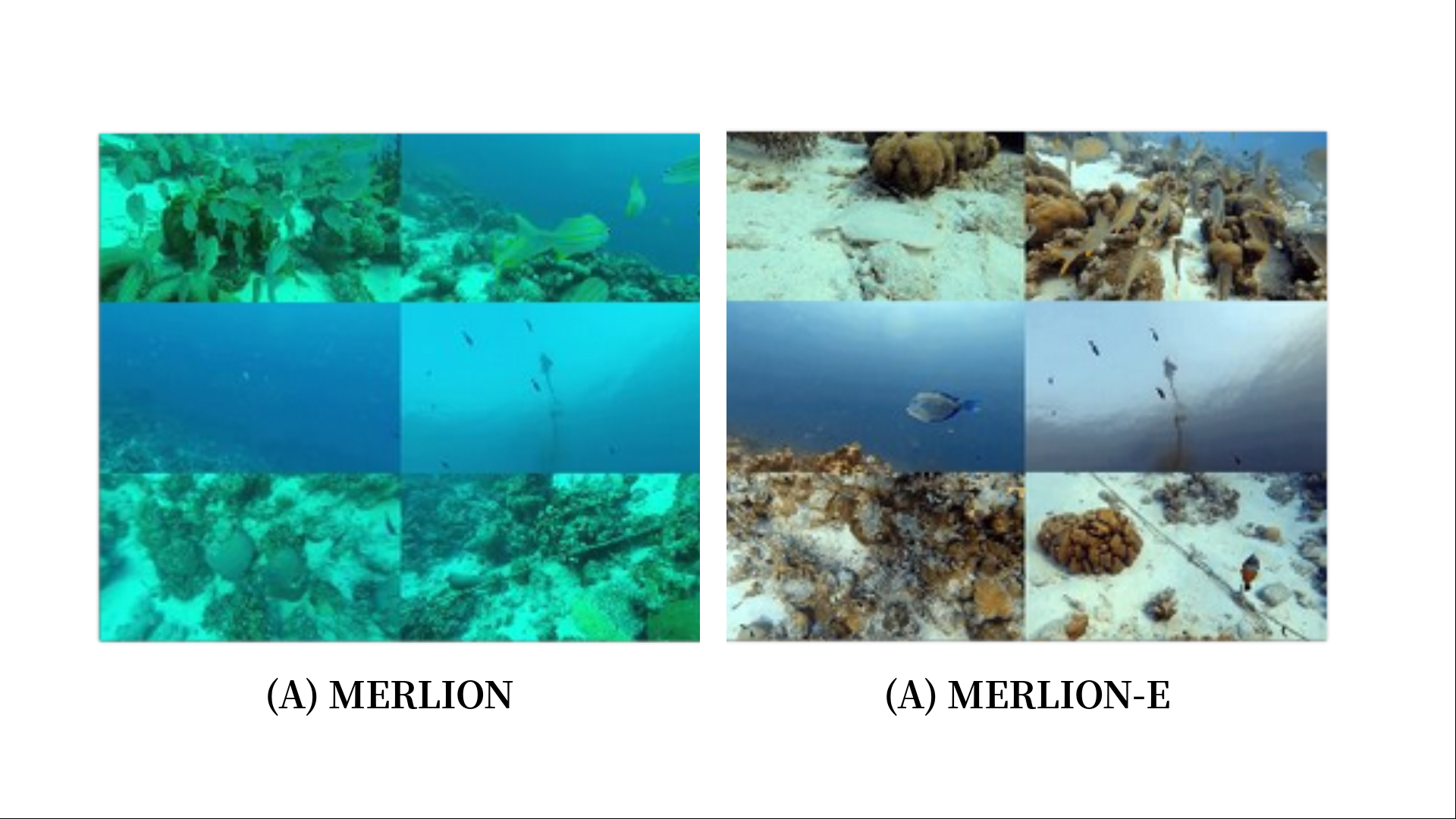} 
    \caption{Semantically aligned and visually enhanced samples obtained using our proposed framework: MERLION [Left] and MERLION-E [Right] with alignment of 48\% and 98\% with human subjects respectively.}
    \label{fig:merlion}
\end{figure}


Advancements in deep learning have enabled tractable solutions for both visual data enhancement and sampling, independently. The challenge lies in integrating them harmoniously and within the confines of AUV computers, while accommodating diverse goals of marine missions, requiring a possibly novel solution. Specifically, we address the following key challenges: (a) delicate balancing between visual data enhancement, semantic sampling and semantic customizability, (b) developing a visual data enhancement algorithm that transforms an unclear image significantly enough to influence sampling, while maximizing computational speed; (c) affording a semantic understanding that is both generalizable and customizable; and (d) developing a standardized approach of evaluation.

Firstly, we address the challenge of balancing between visual data enhancement, semantic sampling and semantic customizability. We choose CLIP \cite{radford2021learning} as an integral factor for addressing the aforementioned challenges. Specifically, CLIP is robust at very low image resolutions which facilitates the reduction of the computation time of visual data enhancement. CLIP allows us to capture the semantic diversity of the scene by using a $512$-vector encoding of the image. This reduced data representation encoding is suitable for integration with our previous work on visual sampling, SON-IS \cite{thengane2024online}, to provide tractable and semantically aligned visual samples.
In addition, CLIP encoding facilitates user customization through natural language text input which allows for semantic customization. 

Secondly, to develop a visual data enhancement algorithm suitable for our sampling task, we need it to be sufficiently transformative to influence the downstream task of sampling. The technique of diffusion is chosen as it produces maximal visual data enhancement. 
However, this is achieved with a trade-off on the computation speed which is solved by using low resolution input image. 
This could significantly degrade the performance of the downstream task of sampling. Due to CLIP's aforementioned robustness, this effect is mitigated. 
We elaborate further on this in Section \ref{sec:AUVImp}. In addition, we employ the selective enhancement strategy explained in Section \ref{sec:MERLION-E}.

Thirdly, the challenge of affording a semantic understanding that is both generalizable and customizable is important, as marine robotics missions can span across subjects from fish to underwater structures. Using the CLIP model and text queries, we customize by filtering out visual frames which fall below a threshold of semantic relevance. The determination of the threshold is discussed in Section \ref{sec:results_and_discussion}. In addition, CLIP is an open-set model which also allows for generalizability across various domains.


Finally, we develop a robust and standardized method to evaluate our proposed system. Specifically, the robustness is achieved with the use of semantically diverse labeled datasets with different degrees of visibility. The standardization is achieved using our semantic scoring validated by human user studies.


In addressing these key challenges, we present a novel framework, MERLION i.e., Marine ExploRation with Language guIded Online iNformative visual sampling and enhancement. Specifically, our contributions include:
\begin{enumerate}[label=\alph*)] 
    \item a novel framework that produces semantically aligned and visually enhanced summaries in real time;
    \item method of validation of our framework using real-world data and user studies; and
    \item an open-source implementation including the evaluation tools.
\end{enumerate} 
This makes MERLION a suitable real-time solution for visual monitoring and exploration on AUVs in murky underwater marine environments. The results of our framework implementation are presented in Fig. \ref{fig:merlion}.


\section{Related Work}
\label{sec:related_work}


We categorize the related work based on the sub-modules of our MERLION framework. This includes (a) semantic feature extraction, (b) informative sampling and (c) image enhancement.

\subsection{Semantic Feature Extraction} 

The CLIP \cite{radford2021learning} model is a semantic feature extractor that integrates two parallel encoders, one for text ($\mathbf{E}_{\text{text}}$) and the other for images ($\mathbf{E}_{\text{image}}$).
Both encoders use feature embeddings with the same dimensionality, facilitating the learning of a unified representation space for both modalities. The model is trained using a contrastive loss, which encourages high feature similarity between image and text embedding belonging to the same image-caption pair, aligning them in a joint feature space. Given a batch of image-text pairs, it maximizes the cosine similarity between matched pairs while minimizing the similarity between unmatched image-text embedding pairs.
Vision-language pre-trained models learn open-set visual concepts via the high-capacity text encoder. This broader semantic space makes learned representations more transferable to downstream tasks, offering impressive zero-shot capabilities. In this work we plan to leverage this generalization capability of CLIP for feature extraction in our visual sampler without additional training and with customization capabilities for the user.

\subsection{Informative Sampling}
There has been much research into the topic of visual informative sampling, related to sampling visual frames in both online and offline manner \cite{thengane2024online}, \cite{girdhar_autonomous_2014}, \cite{inproceedings}, \cite{10.1007/978-3-319-10584-0_33}. In this paper, offline strategy for visual sampling refers to "Video Summarization" in the related literature. Detailed examination of these studies will be provided in the following section.

\subsubsection{Online Sampling}
Girdhar et al.\cite{girdhar_autonomous_2014} introduced ROST (Realtime Online Spatio-temporal Topic modelling) framework, which is an online sampling algorithm that utilizes semantic-agnostic features such as ORB, color, Gabor, and Texton, then input into a topic model which can be used to create online summaries of unique observations using a surprise factor. ROST determines the surprise factor in the generation of samples ("summaries") by comparing between the topic distributions of current and past visual observations in real time, updating the online samples accordingly. Another approach \cite{inproceedings}, Online Visual Summarization, suggests a convolutional Long Short-Term Memory (LSTM) network to encode spatial and temporal features for online sampling. However, it is a supervised learning approach which requires prior information about visual length during its training process. For underwater applications using AUVs, the unsupervised clustering method used in ROST is K-OnlineSummaryUpdate algorithm \cite{girdhar_autonomous_2014}.
This algorithm calculates the surprise score and substitutes an incoming surprising frame for an old one iteratively in real time which helps in maintaining a tractable size of the online summary. We adapt this technique in our proposed MERLION framework for semantic sampling, as we did in SON-IS\cite{thengane2024online}.




\subsubsection{Offline Sampling}
In contrast to online sampling, \cite{10.1007/978-3-319-10584-0_33} and \cite{otani2016video-old} 
utilize semantic features for dynamic informative sampling in an offline manner. They employ a clustering-based sampling technique, which requires the entire dataset beforehand, extracting learning-based features from individual visual segments to produce the visual samples. Despite their effectiveness, these approaches lack real-time adaptability. In \cite{jiang_comprehensive_2019}, \cite{sciencedirectVSUMMMechanism}, \cite{6786125}, \cite{Wei_Ni_Yan_Yu_Yang_Yao_2018}, \cite{sharghi2017queryfocused}, \cite{NIPS2014_0eec27c4} advanced and diverse techniques are used for offline sampling, but are not readily available or infeasible to be used in an online manner. Our proposed work MERLION seeks to tackle the visual sampling problem using similar advanced techniques from other models but tailored for real-time use cases, which is more suitable for AUVs. 

\subsection{Image Enhancement}
\label{sec:image_enha_module}




Image generation models have been recently successfully used for image enhancement. Some of the state-of-the-art techniques include Variational Auto Encoder (VAE)\cite{chira2022image}, Generative Adversarial Networks (GANs) \cite{islam2019fast} or Denoising Diffusion Probabilistic Models (DDPM) \cite{DBLP:journals/corr/abs-2006-11239}. From these models, the diffusion model has provided promising performance for visual data enhancement, especially, for the downstream tasks such as visual sampling.

Standard diffusion models generate images using noise, limiting their effectiveness in enhancing murky underwater images. To address this, \cite{saharia2022image} introduces a conditional diffusion model, incorporating a reference image to guide the enhancement process. Similarly, \cite{2023arXiv230903445T} applies this approach
to the underwater environment which we implement for our visual data enhancement module.
 

\section{methodology}

\begin{figure*}[h]
    \centering
    \includegraphics[width=\linewidth, clip]{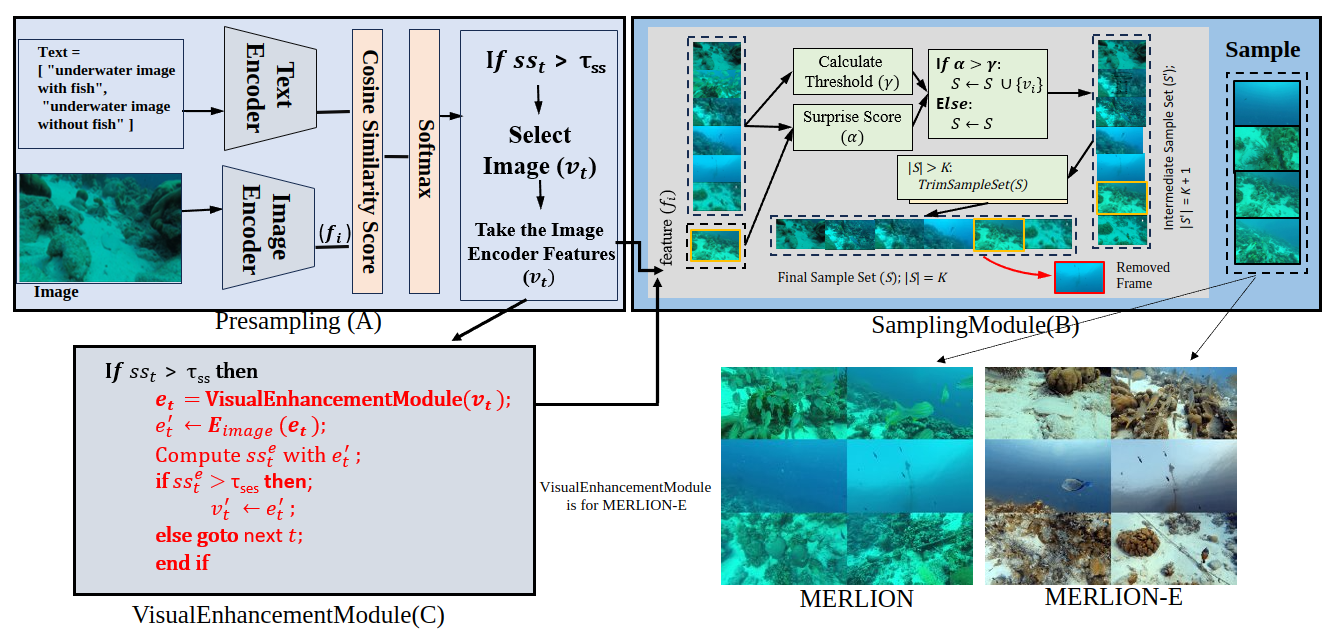} 
    \caption{Overview of the MERLION and MERLION-E framework for underwater visual sampling. The pipeline consists of three main modules: \textit{\bf(A) PresamplingModule}, where text and image encoders compute cosine similarity scores to preselect relevant samples; \textit{\bf(B) SamplingModule}, which refines the selected samples based on a surprise score($\alpha$) and threshold($\gamma$) to form the final sample set; and \textit{\bf(C) VisualEnhancementModule}, used exclusively in MERLION-E, which enhances selected samples before final selection. The bottom right comparision shows the sample outputs from MERLION and MERLION-E, where MERLION-E produces visually enhanced samples.}
    \label{fig:proposed_method}
\end{figure*}

In this section, we introduce our proposed framework MERLION, a semantically aligned online visual data sampler. The overview of our proposed framework is presented in Fig. \ref{fig:proposed_method} and the pseudocode is provided in Algorithm \ref{alg:proposed_method}.
MERLION consists of two main components: a feature extraction (pre-sampling) module based on CLIP and a sampling module (discussed in Algorithm \ref{alg:proposed_method}). Additionally, we present an extension of MERLION designed to enhance murky underwater visual data, denoted as MERLION-E. This extension incorporates an additional image enhancement module (discussed in Section \ref{sec:image_enha_module}) to improve performance in challenging underwater conditions.

\subsubsection{\bf MERLION} 
Given an input visual stream $\mathbf{V} = \{v_t \mid t \in N\}; v_t \in \mathbb{R}^{H \times W \times N}$ with $N$ frames, where $N$ is unknown during the process, and user text queries $\mathbf{q} = \{q_j; j \in M\}$, which includes one positive query and one or more negative queries\footnote{Positive queries are represent desired semantic topics for the summary, while negative queries represent the undesirable ones.}, where $M$ is the number of queries, our method aims to create a subset $\mathcal{S}$ of visual dataset $\mathbf{V}$ of size $K$, where $K \ll N$. 

\textit{\bf Semantic Feature Extraction Module:} We compute features for the user queries by leveraging the pre-trained CLIP \cite{radford2021learning} text encoder $\mathbf{E}_{\text{text}}$, which transforms each query $\mathbf{q}$ into a feature representation $\mathbf{q}^\prime = \mathbf{E}_\text{text}(\mathbf{q})$ of size $\mathbb{R}^{M \times 512}$. At time $t$, as each new visual frame $\mathbf{v}_t$ arrives, we pass it through the CLIP image encoder $\mathbf{E}_{\text{image}}$ to extract the frame feature $\mathbf{v}^\prime_t = \mathbf{E}_\text{image}(\mathbf{\mathbf{v}_t})$ of size $\mathbb{R}^{1 \times 512}$. This allows us to determine which frames contain information relevant to the user queries. To quantify the relevance of each frame to the user queries, we compute a similarity score $\mathtt{ss}_t$ between the current frame $\mathbf{v}_t$ and the user queries $\mathbf{q}$ using softmax cosine similarity: 
\begin{equation}
    \label{eq:cosine_sim}
    \mathtt{ss}_t = \frac{\exp(\text{sim}(\mathbf{v}^\prime_t \cdot \mathbf{q}_1\mathbf{^\prime}))}{\sum_{i=1}^{K} \exp(\text{sim}(\mathbf{v}^\prime_t \cdot \mathbf{q}_i\mathbf{^\prime}))}
\end{equation}

After obtaining a similarity score $\mathtt{ss}_t$, we choose visual frames based on a similarity threshold $\tau_{\mathtt{ss}}$. Frames with similarity scores exceeding this threshold (i.e., $\mathtt{ss}_t > \tau_\mathtt{ss}$) are subsequently passed to the online sampling module.

\textit{\bf Online Informative Sampling Module:} When a frame $\mathbf{v}_t$ matches a text query, we take the image encoder feature vector, denoted as $\mathbf{v}^\prime_t$, and obtain absolute and normalized values before passing it to the online sampler module, as presented in Algorithm \ref{alg:proposed_method}. The output of the online sampler module at $t = K$ is $\mathbf{S} \leftarrow \mathbf{S} \cup \{\mathbf{v}^\prime_t\}$. For each incoming frame  $\mathbf{v}_t$,
we calculate the surprise score $\alpha$ based on the features in the current sample $\mathbf{S}$ and is defined as 
\begin{equation}
   \alpha = \min_i d(v_t, s_i) 
   \label{eq:alpha}
\end{equation} 
\begin{algorithm}[H]
\caption{MERLION: Proposed Methodology}\label{alg:proposed_method}
\begin{algorithmic}
\State \textit{*Content in {\color{citecolor}{red}} color is only used  for MERLION-E method.}
\Input 
    \State $\mathbf{V} = \{v_t \mid t \in N\}; v_t \in \mathbb{R}^{H \times W \times N}$; \Comment{Input visual stream}
    \State $\mathbf{q} = \{q_j \mid j \in M\}$; \Comment{Input text queries}

\Output
 \State $\mathbf{S} \subset \mathbf{V}$; $|\mathbf{S}| = K$; \Comment{Sampled visual frames}
\Require

\State $\mathbf{E}_{\text{image}} \rightarrow$ Image encoder of CLIP;
\State $\mathbf{E}_{\text{text}} \rightarrow$ Text encoder of CLIP;
{\color{citecolor}
\State $\mathtt{VisualEnhancementModule}$ for visual data enhancement;
}

\Method

\State $\mathbf{q}^\prime \gets \mathbf{E}_{\text{text}}(\mathbf{q})$;
\State $\mathbf{S} \leftarrow \{\}$; \Comment{Define empty summary}

\For{$t \in  \{0, \cdots, N\}$}
\State $\mathbf{v}^\prime_t \gets \mathbf{E}_{\text{image}}(\mathbf{v}_t)$;
\If{$t \leq K$}
    \State $\mathbf{S} \leftarrow \mathbf{S} \cup \{\mathbf{v}^\prime_t\}$;
\Else
\State Compute similarity score $\mathtt{ss}_t$ using equation \ref{eq:cosine_sim} with $\mathbf{v}^\prime_t$;
\If{$\mathtt{ss}_t > \tau_{\mathtt{ss}}$}
    {\color{citecolor}
    \State $\mathbf{e}_t = \mathtt{VisualEnhancementModule}(\mathbf{v}_t)$;
    \State $\mathbf{e}^\prime_t \gets \mathbf{E}_{\text{image}}(\mathbf{e}_t)$;
    \State Compute $\mathtt{ss}^{e}_t$ with $\mathbf{e}^\prime_t$;
    \If{$\mathtt{ss}^{e}_t > \tau_{\mathtt{ses}}$};
    \State $\mathbf{v}^\prime_t \gets \mathbf{e}^\prime_t$;
    \Else{ \textbf{goto} next $t$;}
    \EndIf
    }
    \State Compute surprise score $\alpha$ using equation \ref{eq:alpha};
    \State Compute $\gamma$ using equation \ref{eq:gamma};
    \If{$\alpha > \gamma$}
    \State $\mathbf{S} \leftarrow \mathbf{S} \cup \{\mathbf{v}^\prime_t\}$;
    \If {$|\mathbf{S}| > K$} \Comment{Trim Summary}
        \State \textbf{do} $\mathtt{TrimSampleSet}(\mathbf{S})$;
        \State \textbf{while} $|\mathbf{S}| \neq K$;
    \EndIf
    \Else
    \State $\mathbf{S} \gets \mathbf{S}$;

\EndIf
\EndIf
\EndIf
\EndFor
\end{algorithmic}
\end{algorithm}

\newpage
To decide whether to include the current frame in the sample, we compute $\gamma$ as 
\begin{equation}
\gamma = \frac{1}{|\mathcal{S}|} \sum_{i} \min_{j,j \neq i} d(\mathcal{S}_i, \mathcal{S}_j)
\label{eq:gamma}
\end{equation}
If the distance ($\alpha$) between the current frame's features and the sample $\mathbf{S}$ is greater than $\gamma$ then we add the frame in the sample set.
If  $|\mathbf{S}| > \mathbf{K}$ then to maintain the sample's size, less informative frames are removed through a trimming process outlined in \cite{GirdharPhD2014}.

\subsubsection{\textbf{MERLION-E}}
\label{sec:MERLION-E}
MERLION has demonstrated commendable performance, as elaborated in Section \ref{sec:results_and_discussion}. However, its effectiveness diminishes when dealing with input underwater visuals $\mathbf{V}$ that are compromised in murky water or under low light conditions. To address these scenarios, we extended MERLION by incorporating a visual data enhancement module based on the Diffusion model \cite{2023arXiv230903445T}.

Due to the high computational requirements of the diffusion model, we proposed a computationally efficient strategy called ``\textit{selective enhancement sampling (SES)}'' shown in algorithm \ref{alg:proposed_method} in red color. This strategy involves setting the similarity threshold $\tau_\mathtt{ss}$ to a specified value to filter out frames with lower similarity to the text queries. Using this method to remove redundant frames upfront, we significantly reduce computational usage. Subsequently, the remaining frames are passed through the enhancement module, resulting in significant savings of the computational resources. Following the enhancement process, we apply a new threshold, denoted as $\tau_\mathtt{ses}$, to the pre-sampled enhanced images. This threshold is utilized to determine the similarity $\mathtt{ss}^{e}_t$ between the text queries and the pre-sampled enhanced images. If the condition $\mathtt{ss}^{e}_t > \tau_{\mathtt{ses}}$ is met, the remaining sampling process follows the same  procedure as in MERLION.


\section{Results and Discussion}
\label{sec:results_and_discussion}

\subsection{Experimental Setup}
\subsubsection{Dataset and Hyper-parameters}

To assess the efficiency of our method across varied underwater conditions, we curated a diverse set of datasets, each representing a distinct level of visual quality. There are a total of three datasets, the highest visibility dataset is named \texttt{Clear Visibility} \cite{kralendijk}, followed by \texttt{Moderate Visibility} \cite{gopro} which exhibits slight murkiness and lastly \texttt{Low Visibility} \cite{low-vv-dataset}. 
 
To account for the varying degrees of visibility, we set the similarity threshold with respect to both the dataset and the method used. We used lower similarity thresholds $\tau_\mathtt{ss}$ for lower visibility datasets in the MERLION method, and this was empirically set to $40$ for the \texttt{Low Visibility} dataset, $50$ for the \texttt{Moderate Visibility} dataset and $70$ for the \texttt{Clear Visibility} dataset. For the MERLION-E method, the $\tau_\mathtt{ss}$ thresholds was empirically set to $40$ for the \texttt{Low Visibility} dataset and $70$ for the \texttt{Moderate Visibility} dataset. For the MERLION-E method, we also employ the similarity enhancement strategy threshold $\tau_\mathtt{ses}$ of $70$ for all datasets since the visual data has been enhanced. These different thresholds allow our method to adapt to different underwater visibility conditions. 
The thresholds were empirically determined given the variant nature of CLIP's image-text score. In addition, the values are consistent with our hypothesis of proportional relationship between threshold and visual visibility. 


\subsubsection{Baseline}
Given that visual data sampling is a very subjective process,
we rely on human evaluators to validate our framework. To assess visual samples and enable efficient comparison with our approach, human-picked samples from human evaluators were gathered as ground-truth for both original and enhanced visual data. For all three datasets, we conducted a user study with $14$ different human evaluators. The objective was to capture as many unique species of marine animals as possible within six frames. Human evaluators were informed beforehand to sample frames based on specific queries. An example query can be ``\textit{Interesting events related to fish.}''. To simulate an online informative sampling process, the video duration was not shown to the evaluators. 
In addition, they were asked to find representative and unique frames, focusing only on fishes. However, they were permitted to select and save as many potential sample frames while watching the visual sequence. Subsequently, they are required to select the final six frames from the saved potential images without rewinding the visual dataset. We selected a sample size of $6$ as it is sufficient in summarizing both semantics as well as representativeness for our fish datasets. In practice, on AUVs, very large sample sizes may cause computation time and data inefficiencies if not matched with a large semantic and representativeness variety.
For the Clear Visibility dataset, we did not perform enhancement here because the visual quality of the original dataset is already sufficiently high for humans.

\subsubsection{Hardware Compute and AUV Implementation}
\label{sec:AUVImp}
We validated our proposed approach on both a high-grade remote workstation as well as an edge computer on an AUV. 
For the remote workstation, we used RTX $4090$ desktop to process the visual data in real-time. The input image resolution used is $256$x$456$. For the AUV implementation, we tested in murky waters using MERLION-E on a NVIDIA Jetson  Orin AGX 32GB, Seeed Studio H01 variant, running Jetpack 5.12 (Ubuntu 20.04). This computer can be housed within a $5$-inch aluminum tube attached to an upgraded BlueROV2 \cite{marvl-rov}, and with a dedicated $10$Ah battery which is able to run at least for $2$ hours. The input camera feed is sub-sampled to $5$ frames per second and down-scaled to $128$x$128$ resolution (without cropping). With this setup, the algorithm processes an average of $2$-$3$ frames per second on a Orin AGX with comparable performance to the results shown in Section \ref{sec:results_and_discussion}. 


\subsection{Evaluation and Results}
In this section, we present quantitative and qualitative results of MERLION processed using a remote workstation as well as on the edge in an AUV. 

\subsubsection{Remote Workstation}
\begin{figure}[H]
    \centering
    \includegraphics[width=0.5\textwidth]{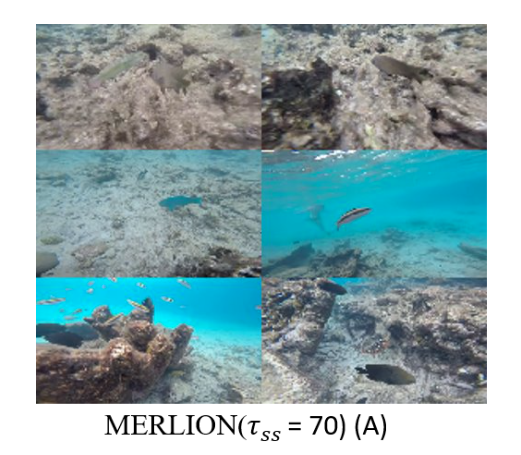} 
    \caption{Qualitative results for a clear Visibility dataset. \cite{kralendijk}} 
    \label{fig:clear_video_sample}
\end{figure}

The quantitative results are presented in Table \ref{tab:quantitative_comparison}. 
In our previous work \cite{thengane2024online}, we proposed an evaluation metric named Semantic Representative Uniqueness Metric (SRUM) which accounts for the semantic and representative scores. This metric calculates the score of the semantics of the automated samples by comparing it with human samples. To do so, we manually label all images in the dataset based on the species of fish identified in the frame.
The Semantic Score rewards a sample frame when its semantic label matches any label from the list of human-selected frames. In such cases, the frame receives a Semantic Score of 1. If no match is found, the frame is assigned a score of 0.
The Representative Score is based on the distance (in time) between the automated sample and matched human samples. And then the final total score is defined as, 
\begin{equation}
    \text{Score} = \alpha \cdot \text{Semantic Score} + (1 - \alpha) \cdot\text{Representative Score}
\end{equation}
Where, $\alpha$ represents the weight of the Semantic Score and $(1 - \alpha)$ is conversely the weight of the Representative Score.
\begin{table*}[!htb]
\caption{Quantitative Comparison}
    \begin{center}
        \label{tab:quantitative_comparison}
        {\scriptsize
        \tabcolsep=0.1cm
        \begin{tabular}{l|ccccc} 
        \toprule
        \rowcolor{purple!20} \textbf{Test Data} & \textbf{MERLION} & \textbf{MERLION-E} & \textbf{ROST} & \textbf{ROST-Enhanced} & \textbf{Human Score}\\
        
        \midrule
        \textbf{Low Visibility} & 0.2446 & 0.4069 & 0.1109 & 0.1384 & 0.6562 \\
        \textbf{Moderate Visibility} & 0.521 & 0.557 & 0.364 & 0.333 & 0.522 \\
        \textbf{Clear Visibility} & 0.4227 & - & 0.3953 & - & 0.5676\\
        \bottomrule
    \end{tabular}}
    \end{center}
    \vspace{-0.5cm}
\end{table*}
The selected samples for \texttt{Clear Visibility} obtained using our proposed approach (MERLION) is shown in Fig. \ref{fig:clear_video_sample}. 

\begin{figure}[h]
    \centering
    \includegraphics[width=\linewidth]{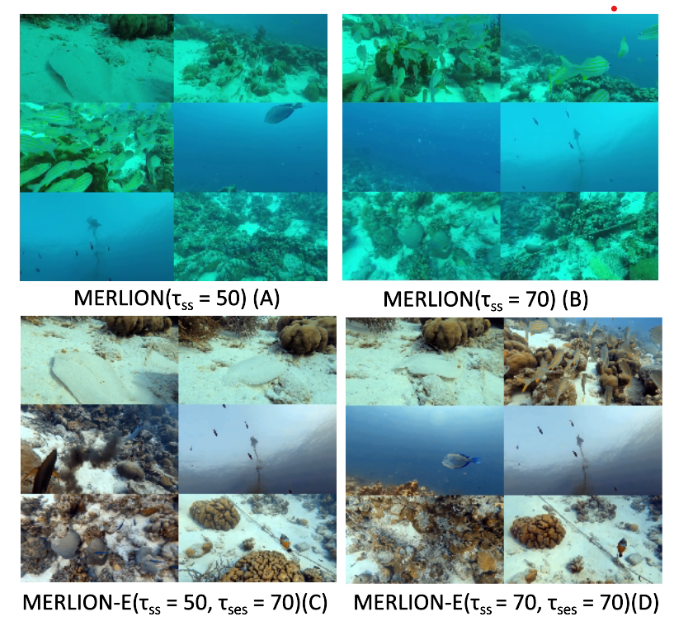} 
    \caption{Qualitative results for \texttt{Moderate Visibility} dataset, sample (A), (B) is for unenhanced visual dataset and sample (C), (D) for the enhanced visual dataset.\cite{gopro} }
    \label{fig:less_murky_sample}
\end{figure}


\begin{figure}[h]
    \centering
    \includegraphics[width=\linewidth]{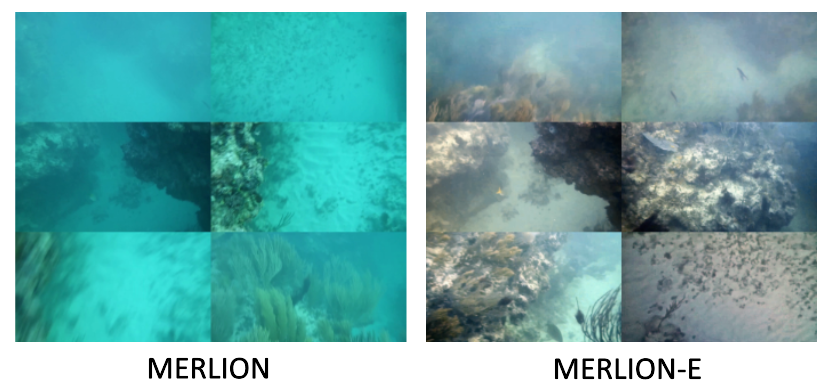} 
    \caption{Qualitative results for the Low Visibility dataset} 
    \label{fig:low_vv_qualitative}
\end{figure}

The quantitative comparison for Low Visibility, Moderate Visibility and Clear Visibility dataset with our proposed method MERLION and MERLION-E are shown in Table \ref{tab:quantitative_comparison}. Human Score is calculated from the average of all human participants' scores against one another (by alternatingly excluding each from the benchmark). ROST scores are obtained from an average of $5$ trials. MERLION and MERLION-E ran deterministically, requiring only $1$ trial each. Across all $3$ datasets, MERLION-E performed the best using the SRUM metric \cite{thengane2024online}. This shows that an enhancement module helps in identifying semantics especially in low visibility conditions. Using our proposed framework of pre-sampling before enhancement, the limitations of a slow enhancement module was also reduced to achieve near real-time performance.

Comparing across datasets for ROST, we see that the murkiness of the visual dataset significantly affects its capabilities in selecting semantically meaningful samples. 
Comparing across datasets for MERLION, it is surprising to see that it performs slightly worse in the Clear Visibility dataset when compared with the Moderate Visibility dataset. This is likely due to the clear dataset having higher overall density of fish species which in turn makes it harder to score both in semantics and representativeness.

When the input visual stream was enhanced, the score does not significantly improve in the ROST-Enhanced column. This is expected as ROST performs informative sampling using the most visually surprising frame which might not correspond to the most semantically surprising frame. The enhancement however, is impactful in the MERLION framework where the score is significantly improved in MERLION-E especially as visibility decreased. This is expected behavior from CLIP, which was trained on cleaned datasets. Thus, a less murky input image is likely to improve CLIP's semantic capabilities as it more closely matches its training distribution. Comparing between MERLION and ROST for the Clear Visibility dataset, both have similar scores with MERLION having a slight edge over ROST. On the other hand, MERLION-E performs significantly better over ROST for both Low Visibility and Moderate Visibility, achieving a score comparable to the Clear Visibility dataset. This signifies that MERLION-E is more robust to murkiness compared to ROST.

The qualitative analysis results on the \texttt{Moderate Visibility} for MERLION and MERLION-E, with varying values of $\tau_\mathtt{ss}$ and $\tau_\mathtt{ses}$, are illustrated in Fig. \ref{fig:less_murky_sample}. The qualitative results for the Low Visibility dataset are illustrated in Fig. \ref{fig:low_vv_qualitative}. We also provide a video at \href{https://www.youtube.com/watch?v=KdlEqcLH9rY}{youtu.be/KdlEqcLH9rY} for more qualitative results. Sub-figures A, B, C, and D correspond to the query ``\texttt{Underwater image with fish.}'' with the opposing query of ``\texttt{Underwater image without fish.}''. The dataset contained corals and rocks as well. It is evident that the model correctly samples images based on the input queries (fish), demonstrating the effectiveness of our approach.

\subsubsection{AUV Edge computing}
Our AUV was deployed off two islands in Singapore with different characteristics. The first island deployment was near the coast and was semantically rich with stable wind and current conditions. In contrast, the second island deployment was in open turbid waters. The results for the latter deployment with MERLION-E on our upgraded BlueROV2 are presented in the following video link: \href{https://www.youtube.com/watch?v=o1hDcecdX5g}{youtu.be/o1hDcecdX5g}. These results showcase the capabilities of MERLION-E in sampling semantically meaningful images such as corals from a highly murky and turbid environment.

\section{Conclusion}
\label{sec:conclusion}

In this work we propose a novel framework, MERLION, that integrates an image-text model for semantically aligning the visual samples to the users' queries, an image enhancement module and an informative sampler. We show that our proposed framework performs the best with quantitative and qualitative results by validating against human evaluated visual sequences obtained from real-world marine environments. MERLION is promising for use on AUVs for the application of long-term vision-based exploration and monitoring in murky waters.


\section*{Acknowledgement}
This project is supported by A*STAR under its RIE2020 Advanced Manufacturing and Engineering (AME) Industry Alignment Fund (Grant No. A20H8a0241) and Google South \& Southeast Asia Research Awards (2024).

\newpage
\bibliographystyle{unsrt}
\bibliography{main}
\end{document}